\documentclass[letterpaper, 10 pt, conference]{ieeeconf}  

\IEEEoverridecommandlockouts                              

\usepackage{graphics} 
\usepackage{graphicx}
\usepackage{booktabs}
\usepackage{amsmath}
\usepackage{epsfig} 
\usepackage{mathptmx} 
\usepackage{times} 
\usepackage{amsmath} 
\usepackage{amssymb}  
\usepackage{multirow}
\usepackage{balance}
\usepackage{pifont} 
\usepackage{makecell}
\usepackage{tabularx} 
\let\labelindent\relax
\usepackage{enumitem}
\usepackage{caption}
\usepackage{float}    
\usepackage{mdframed}

\newcommand{\pmsize}{\scriptsize}
\newcommand{\pmerror}[1]{{\pmsize $\pm$#1}}

\title{\LARGE \bf
Beyond the Majority: Long-tail Imitation Learning for \\ Robotic Manipulation}

\author{Junhong Zhu$^{1*}$, Ji Zhang$^{2*\dagger}$, Jingkuan Song$^{3}$, Lianli Gao$^{1\ddagger}$, Heng Tao Shen$^{3}$
\thanks{$^{1}$Junhong Zhu and Lianli Gao are with University of Electronic Science and Technology of China. $^{2}$Ji Zhang is with Southwest Jiaotong University. $^{3}$Jingkuan Song and Heng Tao Shen are with Tongji University.}%
\thanks{$^{^*}$Equal contribution, $^{\dagger}$Project lead, $^{\ddagger}$Corresponding author.}
}

\begin{document}

\bibliographystyle{unsrt}

\maketitle
\thispagestyle{empty}
\pagestyle{empty}

\begin{abstract}

While generalist robot policies hold significant promise for learning diverse manipulation skills through imitation, their performance is often hindered by the long-tail distribution of training demonstrations. Policies learned on such data, which is heavily skewed towards a few data-rich head tasks, frequently exhibit poor generalization when confronted with the vast number of data-scarce tail tasks.
In this work, we conduct a comprehensive analysis of the pervasive long-tail challenge inherent in policy learning.
Our analysis begins by demonstrating the inefficacy of conventional long-tail learning strategies (e.g., re-sampling) for improving the policy's performance on tail tasks. We then uncover the underlying mechanism for this failure, revealing that data scarcity on tail tasks directly impairs the policy's spatial reasoning capability. To overcome this, we introduce Approaching-Phase Augmentation (APA), a simple yet effective scheme that transfers knowledge from data-rich head tasks to data-scarce tail tasks without requiring external demonstrations.
Extensive experiments in both simulation and real-world manipulation tasks demonstrate the effectiveness of APA. 
Our code and demos are publicly available at: https://mldxy.github.io/Project-VLA-long-tail/.
\end{abstract}

\section{INTRODUCTION}
Recent advances in robotics have been significantly driven by large-scale imitation learning, with generalist robot policies emerging as a dominant paradigm for creating general-purpose agents \cite{sapkota2025vision, ma2024survey, dai2023instructblip, kim2025fine, huang2024embodied}. By training on vast, diverse datasets of human demonstrations, these policies hold the promise of enabling robots to perform a wide range of manipulation tasks from unconstrained natural language instructions. This data-driven approach has already demonstrated remarkable success in acquiring a broad spectrum of manipulation skills  \cite{2024OpenVLA, zhao2025cot, black2410pi0}, paving the way for more flexible and dexterous robotic systems.

\begin{figure}
    \centering
    \includegraphics[width=0.9\linewidth]{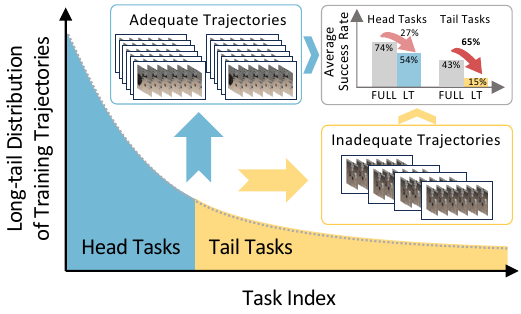}
    \caption{\textbf{Influence of long-tail demonstrations on policy generalization.} Owing to the skewed distribution of training demonstrations, the performance  degradation of generalist robot policies is particularly pronounced for tail tasks relative to those in the head.}
    \label{fig:LT problem}
\end{figure}

However, a critical and commonly disregarded challenge hinders the real-world applicability of these models: the naturally long-tail distribution of demonstration data. In large-scale robotics dataset, a small number of common head tasks (e.g., \texttt{pick up the bowl/plate}) are demonstrated frequently, while the vast majority of tail tasks (e.g., \texttt{put the wine bottle on the rack}) are represented by only a handful of examples \cite{2024OpenVLA, o2024open, liu2023LIBERO}. As we will demonstrate, policies trained on such imbalanced data often perform unreliably on tail tasks \cite{zhao2024ltgc, zhang2023deep}. Fig.~\ref{fig:LT problem} shows that when shifting from a full (or balanced) to a long-tail (or imbalanced) dataset, the relative performance degradation is far more pronounced for tail tasks than for head tasks. This long-tail problem represents a fundamental challenge to developing generalist and reliable robotic agents.

A natural first step to address this issue would be to apply conventional long-tail learning strategies that have been successful in computer vision, such as class re-balancing method or data augmentation techniques \cite{kang2019decoupling,zhang2023deep}. 
However, we find that these methods are largely ineffective for policy learning.
Specifically, our evaluation of re-sampling strategies from \cite{kang2019decoupling} revealed limited efficacy, as they merely duplicate existing data without introducing the variational diversity necessary for generalization.
Furthermore, augmentation techniques like mixup, while successful in classification, are incompatible with robotic control. The linear interpolation of states and actions disregards underlying dynamics, frequently resulting in kinematically infeasible or physically invalid trajectories \cite{kim2013maximum, werner2025llm}. 
This raises the following questions:
\begin{mdframed}
[skipabove=4pt, innertopmargin=4pt, innerbottommargin=4pt]
\textit{What is the core reason for the performance drop of policies trained on long-tail demonstrations? And how can this issue be resolved without relying on external demonstration data?}
\end{mdframed}

To answer these questions, we construct a long-tail imitation learning benchmark based on the LIBERO environment \cite{liu2023LIBERO}, comprising 10 diverse manipulation tasks. Through an in-depth investigation, we find that policies fail on tail tasks primarily due to a degradation of their spatial reasoning capabilities. Because of training data scarcity, the model is unable to learn the precise spatial relationships needed for tail tasks, although it has grasped the general concept from head tasks. Based on this key insight, we introduce Approaching-Phase Augmentation (APA), a simple yet effective scheme designed to improve the spatial reasoning capabilities of robot policies on tail tasks. APA is a self-contained method that leverages demonstrations from head tasks to generate new, high-quality training examples for tail tasks. Crucially, it achieves this without requiring any external demonstration data, making it a practical and efficient solution. We conduct extensive evaluations in both simulation and real-world environments, the achieved results across a total of 16 diverse manipulation tasks demonstrate the effectiveness of our method.

Our contributions in this work are threefold.
\begin{itemize}
    \item We construct a long-tail imitation learning benchmark based on the LIBERO environment, and identify that data scarcity on tail tasks directly impairs the policy’s spatial reasoning capability.
    \item We propose Approaching-Phase Augmentation (APA), a simple and effective method that improves spatial reasoning by transferring knowledge from head tasks to tail tasks without requiring external demonstrations.
    \item Extensive experiments in both simulation and the real-world validate the effectiveness of our APA method.
\end{itemize}

\section{Related works}

\subsection{\textbf{Generalist Robot Policies}}
Recent advances in embodied AI have been driven by robot policies, which integrate perception, language understanding, and control into an end-to-end policy. Pioneering works like RT-1 \cite{rt12022arxiv} introduced the tokenization of robot actions and observations for control within a transformer architecture. RT-2 \cite{zitkovich2023rt} significantly advanced this by reframing robotic control as a vision-language problem, directly transferring web-scale knowledge to novel robotic tasks. Subsequently, Octo \cite{team2024octo} was developed as a generalist, multi-task robot policy with a unified architecture, serving as a versatile foundation model adaptable across diverse robot embodiments. OpenVLA \cite{2024OpenVLA} established a new state-of-the-art by integrating a powerful vision encoder with a modern LLM, setting a new open-source benchmark for imitation learning performance. SmolVLA \cite{shukor2025smolvla} presented an efficient Vision-Language-Action model that lowered the barrier for robotic learning by enabling training and deployment on consumer hardware using community-sourced data. Physical Intelligence company proposed $\pi_0$ \cite{black2410pi0}, prioritizing high-dexterity control with continuous actions generated via flow-matching networks. $\pi_0$-FAST variant \cite{pertsch2025fast} reverted to discrete actions for faster training at the expense of slower inference. Regardless of the model, task success rates are heavily dependent on the quantity and quality of training data.

\subsection{\textbf{Learning from Long-tail Distributions}}
Learning from long-tailed data remains a significant challenge in machine learning, where models are prone to bias towards head classes due to the severely imbalanced data distribution \cite{liu2019large, he2017mask, liu2025survey,zhangji2023channel,zhangji2025reliable}. To address this issue: 1) re-sampling techniques \cite{shi2023re, guo2021long, cui2019class} balance the training data distribution by either oversampling tail-class instances or undersampling head-class instances, thereby reducing the frequency disparity across classes; 2) data augmentation strategies \cite{li2021metasaug, chen2022imagine,chen2024instance} enhance the diversity and volume of tail-class samples through generative or transformation-based methods. Both streams strive to mitigate model bias and improve generalization on tail classes, albeit through distinct mechanisms. 
A more elegant and popular solution is head-to-tail knowledge transfer, which leverages head class data as a reservoir of transferable structure to enhance the learning of tail classes. SMART \cite{zhang2024semantic} transferred the semantic covariance of head classes to enrich the feature patterns of tail classes. \cite{liu2020deep} introduced an embedding-augmentation framework that transferred intra-class angular distributions from head to tail. GLMC \cite{du2023global} mitigated classifier bias by applying a cumulative reweighted loss to soft labels generated from mixed head and tail samples. H2T-FAST \cite{meng2023h2t} performed feature-level style transfer, injecting head-class styles such as textures and color patterns into tail-class content to synthesize augmented features. Copy-Paste \cite{ghiasi2021simple} performed object-level augmentation by pasting tail objects onto diverse backgrounds, often derived from head images, to increase data diversity. 

\section{Diagnosing the Long-tail Failure Mode}
\label{sec:Deconstructing}
To investigate the long-tail challenge in robot policies, we first formalize the problem and construct imbalanced datasets based on the LIBERO \cite{liu2023LIBERO} benchmark. Subsequently, we demonstrate that traditional re-sampling methods are ineffective for VLAs, and then conduct a systematic analysis to identify the root causes of performance degradation.
\subsection{\textbf{Problem formulation}}
A robot policy $\pi_{\theta}$ is able to map multimodal observations and a natural language instruction to a sequence of executable actions for a specified task \cite{2024OpenVLA}. Let the observation at time $t$ be $o_t \in O$ and the instruction be $L$. The observation space typically includes visual input (e.g., RGB images $I_t \in \mathbb{R}^{n \times H \times W \times 3}$, where $n \geq 1$) and may include proprioceptive states (e.g., joint angles). Commonly, the action at time $t$ is $a_t \in A$, $a_t \in \mathbb{R}^7$  represents the end-effector pose, which uses 7-DOF for each arm \cite{liu2025hybridvla, li2024cogact}. Each 7-DOF action includes 3-DOF relative translation offsets ($[\Delta x,\Delta y,\Delta z]\in \mathbb{R}^3$, 3-DOF rotation euler angles( $[Roll, Pitch, Yaw]\in \mathbb{R}^3$), and 1-DOF gripper state (Open/Closed $\in \mathbb{R}^1$). For each demonstration $d$ in the dataset $D$, the trajectory is written as $(\mathbf{o}^{(d)}, \mathbf{a}^{(d)}, L)$ with a sequence of states $\mathbf{o}^{(d)}=(o^{(d)}_{1},\dots,o^{(d)}_{T_i})$ and a corresponding sequence of actions $\mathbf{a}^{(d)}=(a^{(d)}_{1},\dots,a^{(d)}_{T_i})$, where $T_i$ is the length of the trajectory. We follow the standard language-conditioned imitation learning setting to train a policy $\pi_{\theta}$. Its parameters ${\theta}$ are optimized by minimizing:
\begin{equation}
\min_{\theta}\; J(\theta)
= \mathbb{E}_{(\mathbf{o},\mathbf{a},L)\sim D}
\left[ \sum_{t=1}^{T} \mathcal{L}_{\text{action}}\big(\pi_{\theta}(s_t, L),\, a_t\big) \right],
\label{eq:il_objective}
\end{equation}
where $\mathcal{L}_{\text{act}}$ denotes the action loss function (e.g., Mean Squared Error for continuous actions and Cross-Entropy Loss for discrete actions). At inference time, the model takes $(o_t, L)$ as input and outputs $\hat{a}_t=\pi_{\theta}(o_t, L)$.

\subsection{\textbf{A Long-tail Imitation Learning Benchmark}}
\label{data construction}
\begin{figure}
    \centering
    \includegraphics[width=0.9\linewidth]{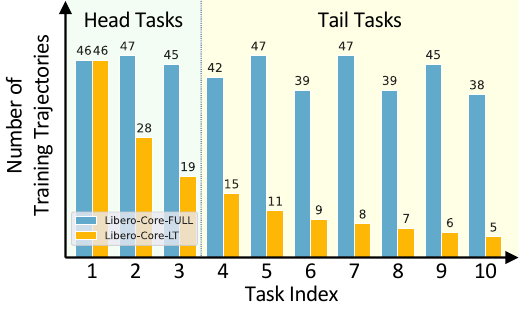}
    \caption{Comparison of training demonstration distribution between the \texttt{LIBERO-Core-FULL} (blue bar) and \texttt{LIBERO-Core-LT} (yellow bar) datasets.}
    \label{fig:distribution of new dataset}
\end{figure}

We introduce two task sets based on the LIBERO environment \cite{liu2023LIBERO}. First, we curate \texttt{LIBERO-Core-FULL} by selecting 10 representative tasks from the LIBERO-Spatial, LIBERO-Object, and LIBERO-Goal splits. Subsequently, we construct \texttt{LIBERO-Core-LT} by generating a long-tailed training set from these 10 tasks. We first order the 10 tasks according to the skill hierarchy of the Open X-Embodiment dataset \cite{o2024open}. Drawing inspiration from prior work in computer vision, such as ImageNet-LT and Places-LT \cite{liu2019large}, We then sample demonstrations for each task using a Pareto distribution \cite{reed2001pareto}, which allocates a varying number of demonstrations to each task. Following the proportional partitioning methodology commonly used in long-tailed recognition, we categorize tasks by their per-task demonstration counts: the top 30\% (3 of 10 tasks) constitute the data-rich head, and the remaining 70\% (7 of 10 tasks) form the data-scarce tail.
The language instructions and the number of training demonstrations for each task are provided in Table~\ref{tab:appendix_tasks}. A comparison of the demonstration distribution between the two datasets is shown in Figure~\ref{fig:distribution of new dataset}.

\begin{table}[t]
\centering
\caption{The language instructions and the number of demonstrations for each task from \texttt{LIBERO-Core-FULL} and \texttt{LIBERO-Core-LT} datasets.}
\label{tab:appendix_tasks}
\begin{tabularx}{\linewidth}{lXc} 
\toprule
\textbf{Task} & \textbf{Language Instruction} & \textbf{(FULL, LT)} \\
\midrule
Task 1 & Pick up the black bowl next to the plate and place it on the plate & (46, 46) \\
Task 2 & Pick up the black bowl next to the cookie box and place it on the plate & (47, 28) \\
Task 3 & Pick up the black bowl on the cookie box and place it on the plate & (45, 19) \\
Task 4 & Pick up the ketchup and place it in the basket & (42, 15) \\
Task 5 & Pick up the alphabet soup and place it in the basket & (47, 11) \\
Task 6 & Push the plate to the front of the stove & (39, \phantom{0}9) \\
Task 7 & Put the bowl on top of the cabinet & (47, \phantom{0}8) \\
Task 8 & Put the cream cheese in the bowl & (39, \phantom{0}7) \\
Task 9 & Put the wine bottle on top of the cabinet & (45, \phantom{0}6) \\
Task 10 & Put the wine bottle on the rack & (38, \phantom{0}5) \\
\bottomrule
\end{tabularx}
\end{table}

\subsection{\textbf{Inefficiency of Re-sampling Methods}}
\label{sec:resampling}
To evaluate conventional long-tail strategies, we fine-tune the miniVLA model \cite{belkhale2024minivla}, pre-trained on the LIBERO-90 dataset, on our \texttt{LIBERO-Core-LT} training set using various re-sampling methods. All other hyperparameters are held constant to isolate the effect of the training data distribution.

Following the methodology of \cite{kang2019decoupling}, we employ a sampling strategy where the probability $p_j$ of sampling a trajectory from task $j$ is given by:
\begin{equation}
\label{eq:sampling}
p_j = \frac{n_j^q}{\sum_{i=1}^{C} n_i^q},
\end{equation}
where $n_j$ is the number of demonstrations for task $j$, and $C$ is the total number of tasks. The hyperparameter $q \in [0,1]$ controls the degree of re-balancing, where a smaller value of $q$ results in a higher sampling probability for tail tasks with scarce training data. We experiment with several values, setting $q \in \{0.75, 0.5, 0.25\}$, to progressively over-sample from tail tasks. Performance is averaged over 30 rollouts across three distinct random seeds. The results are summarized in Table \ref{tab:RS}. As can be observed, these re-sampling strategies yield marginal performance improvements in only a single case. This suggests that simply increasing exposure to tail-task data is ineffective in improving the policy learning performance on tail tasks.
\begin{table}[tbp] 
    \centering
    \caption{Performance of re-sampling methods.} 
    \label{tab:RS}
    \begin{tabular}{lc}
        \toprule
        \textbf{Method} & \textbf{Success Rate (\%)} \\
        \midrule
        Baseline (Original Distribution) & 26.5 \\
        \cmidrule(lr){1-2} 
        Re-sampling ($q = 0.75$) & 25.1 \\
        Re-sampling ($q = 0.50$) & 25.1 \\
        Re-sampling ($q = 0.25$) & 27.1 \\
        \bottomrule
    \end{tabular}
\end{table}
\begin{table*}[t]
\centering
\caption{Phase-wise failure probabilities and success rates on \texttt{LIBERO-Core-FULL} and \texttt{LIBERO-Core-LT} datasets.}
\label{tab:fullvslt}
\begin{tabular*}{\textwidth}{@{\extracolsep{\fill}}cccccccc}
\toprule
\multicolumn{2}{c}{\multirow{2}{*}{\textbf{Task}}} & \multicolumn{3}{c}{\textbf{LIBERO-Core-FULL}} & \multicolumn{3}{c}{\textbf{LIBERO-Core-LT}} \\
\cmidrule(lr){3-5} \cmidrule(lr){6-8}
 &  & $p_{\text{appr}}$ & $p_{\text{exec}}$ & Success Rate (\%) & $p_{\text{appr}}$ & $p_{\text{exec}}$ & Success Rate (\%) \\
\midrule
\multirow{3}{*}{\quad\quad\quad\quad Head} & Task 1 & - & - & 78.9\pmerror{0.15} & - & - & 56.7\pmerror{0.78} \\
 & Task 2 & - & - & 78.9\pmerror{0.70}  & - & - & 60.0\pmerror{0.00}   \\
 & Task 3 & - & - & 63.3\pmerror{0.33} & - & - & 45.6\pmerror{0.15} \\
\midrule
\multirow{7}{*}{\quad\quad\quad\quad Tail} & Task 4 & 0.078 & 0.244 & 67.8\pmerror{0.04} & 0.111 & 0.322 & 56.7\pmerror{0.78} \\
 & Task 5 & 0.022 & 0.289 & 68.9\pmerror{0.04} & 0.322 & 0.533 & 14.4\pmerror{0.48} \\
 & Task 6 & 0.044  & 0.667  & 28.9\pmerror{0.70}  & 0.189 & 0.556 & 25.6\pmerror{0.04} \\
 & Task 7 & 0.100 & 0.378 & 52.2\pmerror{0.04} & 0.933 & 0.067  & 0.0\pmerror{0.00}    \\
 & Task 8 & 0.167 & 0.422  & 41.1\pmerror{0.44} & 0.433   & 0.544 & 2.2\pmerror{0.15}  \\
 & Task 9 & 0.422 & 0.278 & 30.0\pmerror{0.04} & 0.711 & 0.244 & 4.4\pmerror{0.26}  \\
 & Task 10& 0.089  & 0.800 & 11.1\pmerror{0.00}  & 0.411 & 0.589 & 0.0\pmerror{0.00}    \\
\bottomrule
\end{tabular*}
\end{table*}

\subsection{\textbf{Phase-Wise Failure Analysis for Long-Tail Learning}}
In this section, we seek to answer the following question: What is the core reason for the performance drop of
policies trained on long-tail demonstrations?


To this end, we train two separate model instances for comparison. The first is fine-tuned on our fully balanced dataset, \texttt{LIBERO-Core-FULL}, while the second is trained on its long-tailed variant, \texttt{LIBERO-Core-LT}. To ensure a fair comparison, both models use the same architecture, start from the same pre-trained checkpoint, and employ identical hyperparameters, as shown in Section \ref{sec:resampling}.


\textbf{Phase Decoupling.} 
We decouple each task trajectory into two sequential phases. \textit{1) Target Approaching Phase}: encompassing all actions until the robot's end-effector reaches the immediate vicinity of the primary object. \textit{2) Subsequent Execution Phase}: including all actions that follow a successful approach to complete the task.
Let $n_{\text{appr}}$ denote the number of rollouts failing in target approaching phase, $n_{\text{exec}}$ the number of rollouts that pass target approaching phase but fail in subsequent execution phase, and $n_{\text{succ}}$ the number of fully successful rollouts. From these counts, we compute the unconditional failure probability for target approaching phase ($p_{\text{appr}}$) and the conditional failure probability for subsequent execution phase given success in target approaching phase ($p_{\text{exec}}$), as defined in Equation~\ref{eq:pp1} and Equation~\ref{eq:pp2}, respectively. 
\begin{align}
\label{eq:pp1}
p_{\text{appr}} &= p(\text{appr}=\text{fail}) = \frac{n_{\text{appr}}}{n_{\text{appr}} + n_{\text{exec}} + n_{\text{succ}}},
\end{align}

\begin{align}
\label{eq:pp2}
p_{\text{exec}} &= P(\text{exec}=\text{fail} | \text{appr}=\text{succ}) = \frac{n_{\text{exec}}}{n_{\text{exec}} + n_{\text{succ}}}.
\end{align}

\textbf{Phase-wise Analysis.} 
To quantify the impact of the long-tailed data distribution, we employ Relative Risk (RR) \cite{simon2001understanding}, a concept from epidemiology and economics, to measure the increased likelihood of task failure when training on the \texttt{LIBERO-Core-LT} dataset (the exposed group) versus the \texttt{LIBERO-Core-FULL} dataset (the non-exposed group) \cite{andrade2015understanding}. An RR value greater than 100\% indicates an increased risk, a value of 100\% indicates no change, and a value less than 100\% suggests a decreased risk. The RR for every phase of each task is defined in Equation~\ref{eq:rr_task}, where $i$ denotes the $i^{th}$ of the tasks.
\begin{equation}
\label{eq:rr_task}
    RR_{\text{phase}}^{(i)} = \frac{p_{\text{phase,LT}}^{(i)}}{p_{\text{phase,FULL}}^{(i)}}. 
\end{equation}
To mitigate the influence of extreme values, we compute the geometric mean of the relative risk across all tail tasks, which is well suited for averaging ratios and provides a more stable measure of central tendency. This average relative risk, $RR_{\text{phase}}$, is calculated over the set of tail tasks indexed from $i=m$ to $n$:
\begin{equation}
\label{eq:rr}
RR_{\text{phase}} = \left( \prod_{i=m}^{n} RR_{\text{phase}}^{(i)} \right)^{\frac{1}{n-m+1}}.
\end{equation}
In our setting, the tail tasks are indexed from $m=4$ to $n=10$.


We calculated phase-wise failure probabilities, $p_{\text{appr}}$ and $p_{\text{exec}}$ for each tail task, with the results summarized in Table~\ref{tab:fullvslt}. Based on Equation~\ref{eq:rr_task} and Equation~\ref{eq:rr}, we found that the relative risk of failure during the target approaching phase ($RR_{\text{appr}}$) reached 400.89\% across all tail tasks, compared to 164.34\% in the execution phase ($RR_{\text{exec}}$). This indicates that, relative to a policy trained on the full dataset, a policy trained on long-tail data is four times more likely to fail during target approaching. Notably, the target approaching phase depends most critically on spatial reasoning \cite{qu2025spatialvla,zhang2025inspire,zhang2025pcd} between the robot and the object. This finding reveals that data scarcity in tail tasks directly undermines the policy's spatial reasoning capability.

\section{Approaching-Phase Augmentation (APA)}
\begin{figure*}[t] 
    \centering
    \includegraphics[width=0.9\textwidth]{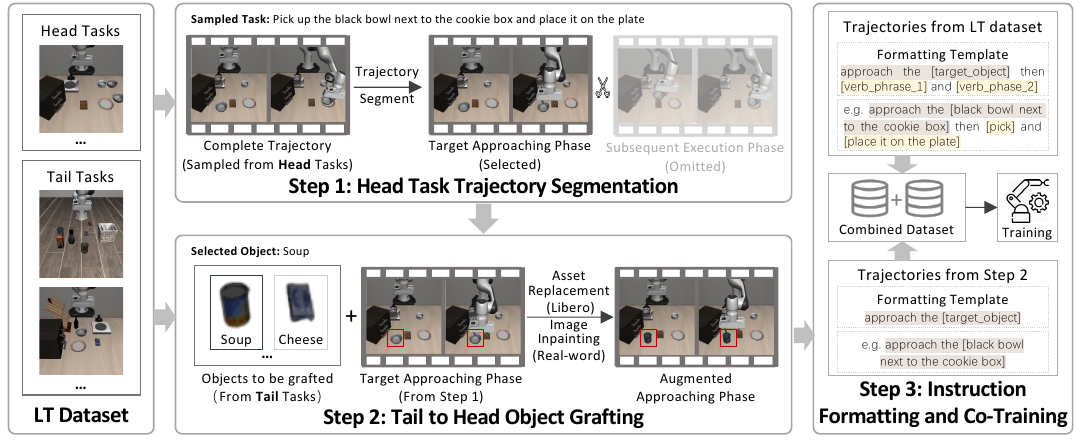} 
    \caption{\textbf{Overview of the Approaching-Phase Augmentation (APA) pipeline.} Our method involves a three-step process: \textbf{(1) Head Task Trajectory Segmentation}, which isolates the target approaching phase from data-rich head tasks; \textbf{(2) Tail to Head Object Grafting}, which creates augmented trajectories using objects from tail tasks; and \textbf{(3) Instruction Formatting and Co-Training}, which formats the corresponding language instructions and then trains the policy on the combined dataset.}
    \label{fig:method_diagram}
\end{figure*}

Motivated by those findings, we propose the Approaching-Phase Augmentation (APA) method,
which leverages demonstrations from head tasks to generate new, high-quality training examples for tail tasks.
The overall process is illustrated in Fig.~\ref{fig:method_diagram}, with the detailed steps outlined below.

\textbf{Step 1: Head Task Trajectory Segmentation.} We begin by randomly selecting a pool of successful execution demonstrations from the data-rich head tasks. For each trajectory, we identify the segment where the robot arm is approaching the target by monitoring its proprioceptive perception (e.g., the gripper's degree of openness or closure). This provides us with a collection of successful approaching-phase demonstrations.

\textbf{Step 2: Tail to Head Object Grafting.} In this step, we generate new training demonstrations by taking a trajectory segment from head tasks and replacing its original object with one from tail tasks. In the simulated LIBERO benchmark, this involves directly replacing the object asset. The new object's initial position is inherited from the source trajectory's original object, while its rotational orientation is specified by the target from tail tasks. The scene is then re-rendered to generate new visual data. For real-world applications, this process can be approximated by identifying the object with YOLOv8 \cite{varghese2024yolov8} and replacing it via image inpainting. Ultimately, this recombination diversifies the training data for the target approaching phase.

\textbf{Step 3: Instruction Formatting and Co-Training.} The final step is to co-train the policy on a composite dataset. To ensure linguistic consistency, we use a pair of corresponding templates to format the language instructions for the original and augmented trajectories, respectively. For our newly augmented trajectories, which only contain the target approaching phase, we generate instructions using the template: \texttt{approach the [target\_object]}. To create a consistent two-phase format across the entire dataset, we modify the instructions for all original demonstrations, instructions for the original demonstrations in source long-tail dataset are modified to \texttt{approach the [target\_object] then [verb\_phrase\_1] and [verb\_phrase\_2]}. The model is then trained on this combined, linguistically consistent dataset.

The primary advantage of APA is its targeted efficiency. By focusing exclusively on augmenting the critical target approaching phase, our method bypasses the complexity of generating full, physically plausible manipulation sequences. Unlike general inpainting methods requiring consistent object properties (e.g., size, mass) for interaction, our approach is unconstrained by these physical requirements. This simple yet effective intervention is possible precisely because it targets the key bottleneck we identified in Section~\ref{sec:Deconstructing}.

\begin{figure}[tbp]
    \centering
    \includegraphics[width=0.88\linewidth]{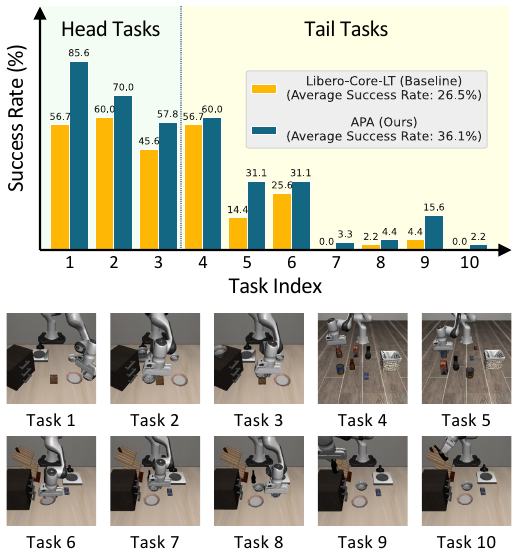}
    \caption{Effectiveness of our proposed APA method on \texttt{LIBERO-Core-LT} dataset.}
    \label{fig:results_barchart}
\end{figure}

\begin{figure*}[t]
    \centering
    \includegraphics[width=0.8\textwidth]{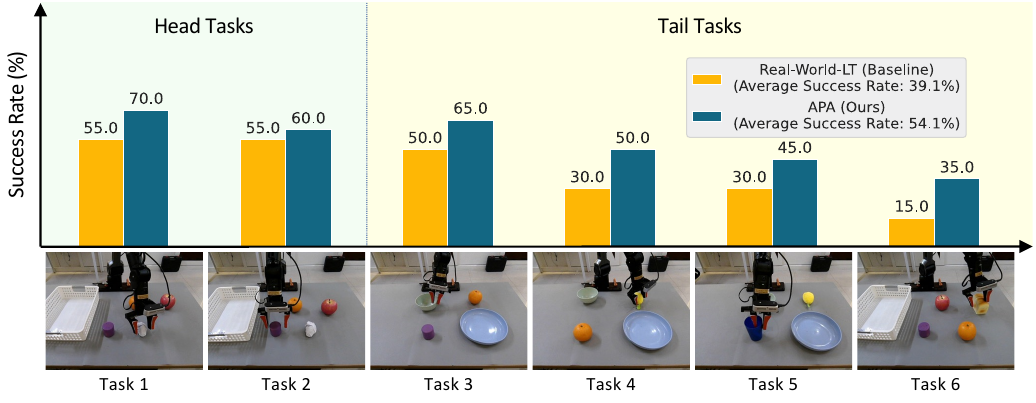} 
    \caption{Effectiveness of our proposed APA method on real-word long-tail tasks.}
    \label{fig:results_realworld}
\end{figure*}

\section{EXPERIMENTS}
In this section, we conduct extensive experiments in both simulation and real-world environments to evaluate the effectiveness of the proposed APA method. 
\subsection{\textbf{Simulation Experiments}}

\subsubsection{\textbf{Experimental Setup}}
Our experiments are conducted on the \texttt{LIBERO-Core-LT} dataset constructed in Section \ref{data construction}. We also employ the state-of-the-art miniVLA as our testbed model, which was pretrained on the LIBERO-90 dataset. To evaluate our method, we fine-tune this checkpoint on two different dataset configurations:
\begin{itemize}
\item \textbf{Baseline:} The model is fine-tuned directly on the original, imbalanced \texttt{LIBERO-Core-LT} dataset.
\item \textbf{APA (Ours):} The model is fine-tuned on the \texttt{LIBERO-Core-LT} dataset augmented with demonstrations generated by our proposed method.
\end{itemize}
Both models were fine-tuned for 36 epochs. Each result was averaged over three random seeds from 30 evaluation rollouts.

\subsubsection{\textbf{Results and Analysis}}
The results, summarized in Fig. \ref{fig:results_barchart}, demonstrate the effectiveness of APA. Our method boosts the average success rate from 26.5\% to 36.1\%, a substantial relative improvement of 36.2\%. 
Crucially, our approach improves performance on data-scarce tail tasks while simultaneously boosting results on data-rich head tasks.
We attribute this positive-sum outcome to our augmentation strategy, which leverages demonstrations from the head as a source. This process appears to provide diverse data for the tail while also reinforcing the learned policies for the head tasks themselves. To confirm that this efficacy is independent of the task distribution, we evaluated APA on a randomly shuffled dataset—an experiment detailed in Appendix \ref{appendix:Shuffled} that underscores the robustness of our approach.

\subsection{\textbf{Ablation Studies}}
We conduct two ablation studies to analyze the key aspects of our APA framework. First, we analyze the individual impact of our core components, Cross-Task Object Grafting and Instruction Formatting. Second, we investigate the effect of varying the number of augmented demonstrations.

\subsubsection{\textbf{Component Analysis}} Our component analysis reveals that the core components of our method are ineffective in isolation and rely on a strong synergy. As detailed in Table \ref{tab:ablation_component}, ablating either the instruction formatting or the trajectory augmentation yields no significant improvement over the baseline. This is because the policy cannot ground the new language instructions without corresponding visual examples, nor can it effectively utilize the new visual data without an explicit linguistic guide. These findings therefore confirm that the instruction formatting and augmentation components are deeply intertwined and codependent, making both essential for achieving optimal performance.
\begin{table}[t]
\centering
\caption{Ablation study on the designed components.}
\label{tab:ablation_component}
\begin{tabular}{lccc}
\toprule
\textbf{Setting} & \textbf{Formatting} & \textbf{Augmentation} & \textbf{Success Rate (\%)} \\
\midrule
Baseline & \ding{55} & \ding{55} & 26.5 \\
+ Formatting & \checkmark & \ding{55} & 26.0 \\
+ Augmentation & \ding{55} & \checkmark & 26.9 \\
\midrule
\textbf{Ours (Full)} & \textbf{\checkmark} & \textbf{\checkmark} & \textbf{36.1} \\
\bottomrule
\end{tabular}
\end{table}

\subsubsection{\textbf{Effect of the Number of Augmented Demonstrations}} We analyze the impact of the number of augmented demonstrations per task added to the training set. Table \ref{tab:ablation_number} shows that while adding these demonstrations is beneficial, the performance gains are not monotonic, and the success rate begins to decline after peaking. We hypothesize that this decline occurs because an excessive number of augmented demonstrations, which only cover the initial target approaching phase, may skew the training data distribution and impact the overall success rate.

\begin{table}[tbp]
\centering
\caption{Impact of the num of augmented demonstrations.}
\label{tab:ablation_number}
\begin{tabular}{ccccc}
\toprule
\textbf{Number of Augmented Demons.} & 0 & 3 & 6 & 9  \\
\midrule
\textbf{Success Rate (\%)} & 26.5 & 28.8 & \textbf{36.1} & 32.0  \\
\bottomrule
\end{tabular}
\end{table}

\subsection{\textbf{Real-world Experiments}}
\subsubsection{\textbf{Experimental Setup}}
To evaluate performance in a physical setting, we design a new long-tail benchmark, named \texttt{Real-World-LT}. This dataset comprises 6 real-world manipulation tasks, for which we collected demonstrations following a Pareto distribution to create a long-tail structure \cite{liu2019large, reed2001pareto}. Further details on the tasks and trajectory counts are provided in the Appendix \ref{appendix:Real-Word}. For the policy, we select the $\pi_0$ model, which was pretrained on the real-world data from OXE \cite{o2024open}. This choice parallels our simulation setup (which used the virtual-pretrained miniVLA), ensuring that we use a base model appropriately matched to its domain. We then evaluate and compare two finetuning approaches: \textbf{Baseline:} The policy is fine-tuned directly on \texttt{Real-World-LT} dataset. \textbf{APA (Ours):} The policy is fine-tuned on the same dataset augmented with APA.
During evaluation, we conduct 20 rollouts for each task, randomizing the configurations and orientations of objects in each trial to ensure a robust assessment of performance. All Our real-world experiments are conducted on an AGILEX PIPER 6-DOF robot arm equipped with a 1-DOF gripper.

\subsubsection{\textbf{Results and Analysis}}
Our real-world experimental results, summarized in Fig.~\ref{fig:results_realworld}, validate the effectiveness of our APA method on a physical robot. Overall, APA achieves an average success rate of 54.1\%, a significant improvement over the baseline's 39.1\%, which corresponds to a 38.4\% relative gain. Consistent with our simulation findings, APA substantially improves the success rate of tail tasks without compromising performance on the data-rich head tasks. In fact, it enhances performance across the entire distribution, demonstrating its comprehensive benefits. Furthermore, these results provide real-world evidence for our initial hypothesis from simulation: that the performance degradation on tail tasks is disproportionately concentrated in the initial object approaching phase, which heavily relies on the model's spatial reasoning capability. The success of our method confirms that by specifically targeting this bottleneck, APA offers a practical and robust solution for mitigating long-tail challenges in robotic manipulation.

\section{CONCLUSIONS}

This paper presents a comprehensive analysis of the long-tail challenge in policy learning. Our in-depth analysis reveals that the primary failure mode of the policy on tail tasks is a degradation of the spatial reasoning capacity. Based on this key insight, we introduce Approaching-Phase Augmentation (APA), a novel and effective method that generates high-quality, targeted training data by grafting tail task objects onto successful approach trajectories from head tasks. Through extensive experiments on both simulated and real-world benchmarks, we show that our APA method significantly boosts the performance on tail tasks without compromising head task success. 




\section*{APPENDIX}

\begin{table}[t]
\centering
\caption{The number of demonstrations for each task from \texttt{LIBERO-Core-LT-Shuffled} dataset.}
\label{tab:appendix_tasks_shuffled}
\begin{tabularx}{\linewidth}{lXc} 
\toprule
\textbf{Task} & \textbf{Language Instruction} & \textbf{Count} \\
\midrule
Task 1 & Pick up the ketchup and place it in the basket & 42 \\
Task 2 & Pick up the black bowl next to the plate and place it on the plate & 29 \\
Task 3 & Pick up the alphabet soup and place it in the basket & 21 \\
Task 4 & Push the plate to the front of the stove & 15 \\
Task 5 & Put the wine bottle on the rack & 12 \\
Task 6 & Pick up the black bowl next to the cookie box and place it on the plate & 10 \\
Task 7 & Pick up the black bowl on the cookie box and place it on the plate & 8 \\
Task 8 & Put the wine bottle on top of the cabinet & 7 \\
Task 9 & Put the cream cheese in the bowl & 6 \\
Task 10 & Put the bowl on top of the cabinet & 5 \\
\bottomrule
\end{tabularx}
\end{table}

\subsection{\textbf{Robustness of APA to Different Task Distributions}}
\begin{figure}[tbp]
    \centering
    \includegraphics[width=0.82\linewidth]{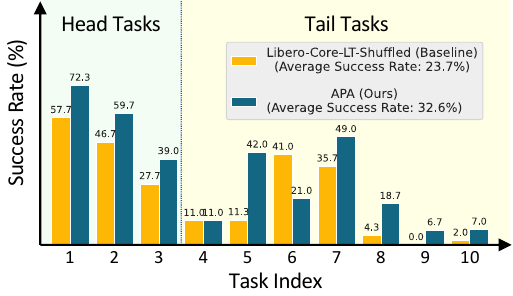}
    \caption{Effectiveness of our proposed APA method on \texttt{LIBERO-Core-LT-Shuffled} dataset.}
    \label{fig:shuffle}
\end{figure}

To verify that the efficacy of our method is not contingent on the specific tasks designated as head in the original long-tail distribution, we evaluate its performance on a new, randomly generated task ordering. We create this new distribution by first shuffling the 10 core tasks and then re-applying Pareto sampling to construct a new long-tailed training set, which we name \texttt{LIBERO-Core-LT-Shuffled}. This procedure effectively reassigns which tasks are data-rich (head) and which are data-scarce (tail), with the new order detailed in Table~\ref{tab:appendix_tasks_shuffled}. The results in Fig.~\ref{fig:shuffle} demonstrate that our approach still outperforms the baseline under this new distribution. This confirms that the effectiveness of APA is both robust and generalizable.

\label{appendix:Shuffled}

\subsection{\textbf{Task Details for the Real-World-LT Dataset}}
\label{appendix:Real-Word}
The details of the six real-world manipulation tasks in \texttt{Real-World-LT} dataset are presented in Table \ref{tab:my_tasks}. 
\begin{table}[tbp]
\centering
\caption{The number of demonstrations for each task from \texttt{Real-World-LT} dataset.}
\label{tab:my_tasks}
\begin{tabularx}{\linewidth}{lXc}
\toprule
\textbf{Task} & \textbf{Language Instruction} & \textbf{Count} \\
\midrule
Task 1 & Pick up the spitball and place it in the basket & 20 \\
Task 2 & Pick up the cylinder and place it in the basket & 13 \\
Task 3 & Put the bowl on the plate & 9 \\
Task 4 & Put the lemon on the plate & 6 \\
Task 5 & Put the cup on the plate & 5 \\
Task 6 & Pick up the bread and place it in the basket & 4 \\
\bottomrule
\end{tabularx}
\end{table}

\section*{ACKNOWLEDGMENT}
This work is supported by the National Natural Science Foundation of China (No.U23A20315, No.62506310, No.62425208), the China Postdoctoral Science Foundation (No.2025M781517), the Postdoctoral Fellowship Program of CPSF (No.GZB20240625), and the Natural Science Foundation of Sichuan Province (No.2026NSFSC1479).


\bibliography{reference}
\end{document}